\def\year{2017}\relax
\documentclass[letterpaper]{article} 
\usepackage{aaai17}  
\usepackage{times}  
\usepackage{helvet}  
\usepackage{courier}  
\usepackage{graphicx}  
\usepackage{ifxetex}
\usepackage{ifluatex}

\ifxetex
  \usepackage[ios,font=seguiemj.ttf]{emoji}
  \usepackage{fontspec}
  
\else
  \ifluatex
    \usepackage[ios,font=Symbola_hint.ttf]{emoji}
    \usepackage{fontspec}
    
  \else
    \usepackage[T1]{fontenc}
    \usepackage[utf8]{inputenc}
    \usepackage[ios]{emoji}
  \fi
\fi
\usepackage{url}

\begin{document}
%

\title{EmojiNet: An Open Service and API for Emoji Sense Discovery}

\author{Sanjaya Wijeratne \and Lakshika Balasuriya \and Amit Sheth \and Derek Doran \\ Kno.e.sis Center, Wright State University\\
Dayton, Ohio, USA\\
\{sanjaya, lakshika, amit, derek\}@knoesis.org}

\maketitle

\begin{abstract}

This paper presents the release of EmojiNet, the largest machine-readable emoji sense inventory that links Unicode emoji representations to their English meanings extracted from the Web. EmojiNet is a dataset consisting of: (i) 12,904 sense labels over 2,389 emoji, which were extracted from the web and linked to machine-readable sense definitions seen in BabelNet; (ii) context words associated with each emoji sense, which are inferred through word embedding models trained over Google News corpus and a Twitter message corpus for each emoji sense definition; and (iii) recognizing discrepancies in the presentation of emoji on different platforms, specification of the most likely platform-based emoji sense for a selected set of emoji. The dataset is hosted as an open service with a REST API and is available at \url{http://emojinet.knoesis.org/}. The development of this dataset, evaluation of its quality, and its applications including emoji sense disambiguation and emoji sense similarity are discussed. 

\end{abstract}

\section{Introduction and Motivation} \label{sec:intro}

With the rise of social media, pictographs, better known as `emoji', have become an extremely popular form of communication. Their popularity may be explained by the typical short text format of social media, with emoji able to express rich content in a single character. Emoji are also a powerful way to express emotions or a hard to write, subtle notion effectively~\cite{kelly2015characterising}. For example, emoji are used by many Internet users, irrespective of their age. Emogi, an Internet marketing firm reports that over 92\% of all online users have used emoji\footnote{\url{https://goo.gl/C5ioVO}}. They further report that emoji use is not simply a millennial fad, as over 65\% of frequent and 28\% of occasional Internet users over the age of 35 use emoji. Creators of the SwiftKey Keyboard for mobile devices report that they process 6 billion messages per day that contain emoji~\cite{swiftkeymost}. Moreover, business organizations have adopted and now accept the use of emoji in professional communication. For example, Appboy, an Internet marketing company, reports that there has been a 777\% year-over-year increase and 20\% month-over-month increase in emoji usage for marketing campaigns by business organizations in 2016\footnote{\url{https://goo.gl/ttxyP1}}. These statistics leave little doubt that emoji are a significant and important aspect of electronic communication across the world.

In the same way that natural language is processed with sophisticated machine learning techniques and technologies~\cite{manning1999foundations} for many important applications, including text similarity~\cite{gomaa2013survey} and word sense disambiguation~\cite{navigli2009word}, so too should emoji be subject to evaluation. Yet the graphical nature of emoji, the fact that (the same) emoji may be used in different contexts to express different senses, and the fact that emoji are used in all languages over the world make it especially difficult to apply traditional NLP techniques to them~\cite{miller2016blissfully,barbieri2016cosmopolitan}. Indeed, when emoji were first introduced, they were defined with no rigid semantics attached, which allowed people to develop their own use and interpretation\footnote{\url{https://goo.gl/ztqjC2}}. Thus, similar to words, emoji can take on different meanings depending on context and part-of-speech (POS)~\cite{emojinet}.

We previously proposed and released a prototype system that maps emoji to their set of possible meanings or {\em senses}~\cite{emojinet}. A sense is defined as a combination of a word (e.g. \texttt{laugh}), its POS tag (e.g. \texttt{noun}), and its definition in a message context or gloss (e.g. \texttt{Produce laughter}). The system was constructed with an eye towards solving the {\em emoji sense disambiguation} problem, which is the ability to identify the meaning of an emoji in the context of a message in a computational manner. The initial system was able to map 845 different emoji symbols (35\% of all emoji supported by the Unicode Consortium) to a set of 3,206 different senses~\cite{emojinet}. However, it did not provide information about all emoji supported by the Unicode Consortium\footnote{\url{https://goo.gl/lo3z1E}}, instead relying on a crowdsourced database (The Emoji Dictionary) with strict rules about deciding when to attach a sense to an emoji. The sense definitions (glosses) available in this system were extracted from BabelNet~\cite{navigli2010babelnet}, which is the most comprehensive machine-readable dictionary available to-date. Those sense definitions are often short (15 words on average) and are based on well written text. Prior research on NLP suggests that the accuracy of the sense disambiguation tasks can be improved by incorporating more context words~\cite{vasilescu2004evaluating} and NLP tools trained on well-formed text might not work well with social media text due to language variations~\cite{ritter2011named}. Thus, it is important for an emoji sense inventory to have access to context words retrieved from social media data streams such as Twitter in addition to well-formed text extracted from BabelNet sense definitions. Moreover, noting past work by Miller {\em{et al.}} on platform-dependent emoji renderings and interpretations~\cite{miller2016blissfully}, our preliminary system did not capture the significant influence a device or interface (e.g. \emoji{1F64C} on Apple iMessage vs. \emoji{1F64X} on Google Messages) has on the intended sense of an emoji. It also did not provide an API for others to access the dataset.

This paper presents the release of EmojiNet, an open service and public API that substantially extends and addresses the limitations of the prototype system discussed above. The service enables researchers and practitioners to query an extensive database of emoji senses, and enables the potential integration of emoji with practical and theoretical NLP analyses. EmojiNet attaches 12,904 sense definitions to over 2,389 emoji, along with data about the relevance of a sense to the platform it is read on for a selected set of emoji. The set of sense definitions extracted from BabelNet for each emoji are strengthened with context words learned from word embedding models from corpuses of Google News articles and Twitter messages. The paper details the architecture of EmojiNet, including its integration with other web resources and the process of disambiguating emoji senses using the contexts learned over the word embedding models. It then discusses the extent of the EmojiNet emoji sense database, the format and metadata stored in it, and provides examples of its use in two use-cases; emoji sense disambiguation and emoji similarity. The paper also gives an evaluation of the quality of emoji pictograph mapping, the quality of the BabelNet sense extraction process, and a qualitative user study using Amazon Mechanical Turk to determine the overall quality of the sense matchings to emoji and the platform it may be rendered on for a set of 40 emoji.

This paper is organized as follows. Section~\ref{sec:rr} discusses the related literature and positions how this work differs from other related works. Section~\ref{sec:dc} discusses how an emoji is modeled in the EmojiNet sense inventory and the techniques adopted to create it. Section~\ref{sec:extend} describes how we further extend the capabilities of EmojiNet as a sense inventory using word embedding models. Section~\ref{sec:eval} reports how we evaluated the processes used to create EmojiNet and the accuracy of the resource. Section~\ref{sec:usecase} discusses two use-cases of EmojiNet and Section~\ref{sec:con} concludes the work reported while discussing the future work planned.

\section{Background and Related Work} \label{sec:rr}

Although emoji was introduced two decades ago, research on this communication form remains limited~\cite{miller2016blissfully}. This may be because emoji were not supported as a standard by The Unicode Consortium until 2009~\cite{unicodeimoji}, which finally enabled its adoption onto mobile platforms~\cite{swiftkeymost}. Early research into emoji focused on understanding the role of emoji in computer mediated communication. Cramer {\em{et al.}} studied the sender-intended functionality of emoji using a group of 228 individuals who used emoji in text messages~\cite{cramer2016sender}. They reported on functional differences in emoji use and showed that the social and linguistic functions of emoji are complex. Kelly {\em{et al.}} reported that people who are in close relationships use emoji as a way of maintaining conversational connections in a playful manner~\cite{kelly2015characterising}. Pavalanathan {\em{et al.}} studied how emoji compete with ASCII-based non-standard orthographies, including emoticons, when it comes to communicating paralinguistic content on social media~\cite{pavalanathan2016more}. They reported that Twitter users prefer emoji over emoticons, and users who adopt emoji tend to use standard English words at an increased rate after emoji adoption. Others have used features extracted from emoji usage in sentiment analysis~\cite{novak2015sentiment}, emotion analysis~\cite{wang2012harnessing}, and Twitter profile classification~\cite{lakshikagang,gangwordembeddings} problems.

Past work on understanding emoji senses by Miller {\em{et al.}} focused on how the sentiment and semantics of emoji differ when the same emoji is displayed on multiple platforms as vendors can design their own emoji image to display~\cite{miller2016blissfully}. Tigwell {\em{et al.}} showed how emoji misunderstanding can happen due to platform-specific designs~\cite{tigwell2016oh}. Barbieri {\em{et al.}} studied emoji meanings using word embeddings~\cite{DBLP:journals/corr/abs-1301-3781} learned over a tweet corpus and used the learned word embeddings to calculate the functional and topical similarity between emoji~\cite{barbieri2016does}. Eisner {\em{et al.}} used emoji descriptions available on Unicode.org to learn emoji meanings~\cite{eisner2016emoji2vec} and showed that their emoji representation model could outperform the Barbieri {\em{et al.}}'s model in a sentiment analysis task.

Work on building resources that enable the natural language interpretation of emoji is at a very early stage.  Several web resources list emoji senses either as keywords or sense labels, which is defined as a \texttt{word(PoS tag)} pair such as \texttt{laugh(noun)}. Sense labels can be helpful for developing emoji sense inventories. For example, The Unicode Consortium\footnote{\url{https://goo.gl/lo3z1E}} and EmojiLib\footnote{\url{https://goo.gl/2nJIHu}} provide lists of keywords that could act as the intended meanings for emoji. The Emoji Dictionary lists sense labels for emoji meanings that are collected via crowdsourcing. However, none of these web resources can serve as machine-readable sense inventories due to the limitations in their system designs, including not providing enough training examples for a computer program to understand how an emoji should be used in a message context~\cite{emojinet}. Therefore, simply scraping those websites to extract emoji sense labels alone cannot help to build emoji sense inventories out of them. The sense labels need to be linked with machine processable dictionaries such as BabelNet to extract message contexts for them. The Emoji Dictionary contains more valuable emoji meanings compared to what the Unicode Consortium website and EmojiLib have to offer, but The Emoji Dictionary does not list unicode representations of emoji, which makes it difficult to be directly consumed by a computer program.

In an initial experiment, we integrated four openly available emoji resources on the web and link sense labels extracted from The Emoji Dictionary with BabelNet to create a new sense inventory~\cite{emojinet}. The  focus of this work was on the actual construction of a sense inventory, rather than ensuring that it is complete and provides comprehensive sense information about emoji. Hence, our initial system only covers emoji for which we could extract emoji sense definitions from The Emoji Dictionary. This system only provides emoji sense definitions for 845 out of the 2,389 emoji supported by the Unicode Consortium (only 35\% of all supported emoji). Moreover, the sense definitions available in this system are short and extracted from well-formed text, which is not suitable for building tools for NLP tasks over corpuses of non-standard and ill-formatted text with heavy emoji use~\cite{ritter2011named}. Finally, this system cannot be used to learn platform-specific (or vendor-specific) emoji senses as pointed out by~\cite{miller2016blissfully} and the dataset used to build the system is not available publicly to download. Our new release address the above limitations, and includes a REST API for accessing a complete dataset of emoji senses.

\section{Emoji Modeling and Dataset Creation} \label{sec:dc}

EmojiNet exposes a dataset of emoji. Each emoji is represented as a nonuple representing its sense and other metadata. Let $E$ be the set of all emoji in EmojiNet. For each emoji $e_i \in E$, EmojiNet records the nonuple $e_i = (u_i, n_i, c_i, d_i, K_i, I_i, R_i, H_i, S_i)$, where $u_i$ is the Unicode representation of $e_i$, $n_i$ is the name of $e_i$, $c_i$ is the short code of $e_i$, $d_i$ is a description of $e_i$, $K_i$ is the set of keywords that describe intended meanings attached to $e_i$, $I_i$ is the set of images that are used in different rendering platforms, $R_i$ is the set of related emoji extracted for $e_i$, $H_i$ is the set of categories that $e_i$ belongs to, and $S_i$ is the set of different senses in which $e_i$ can be used within a sentence.

\begin{figure}
\centering
\includegraphics[width=1.0\linewidth]{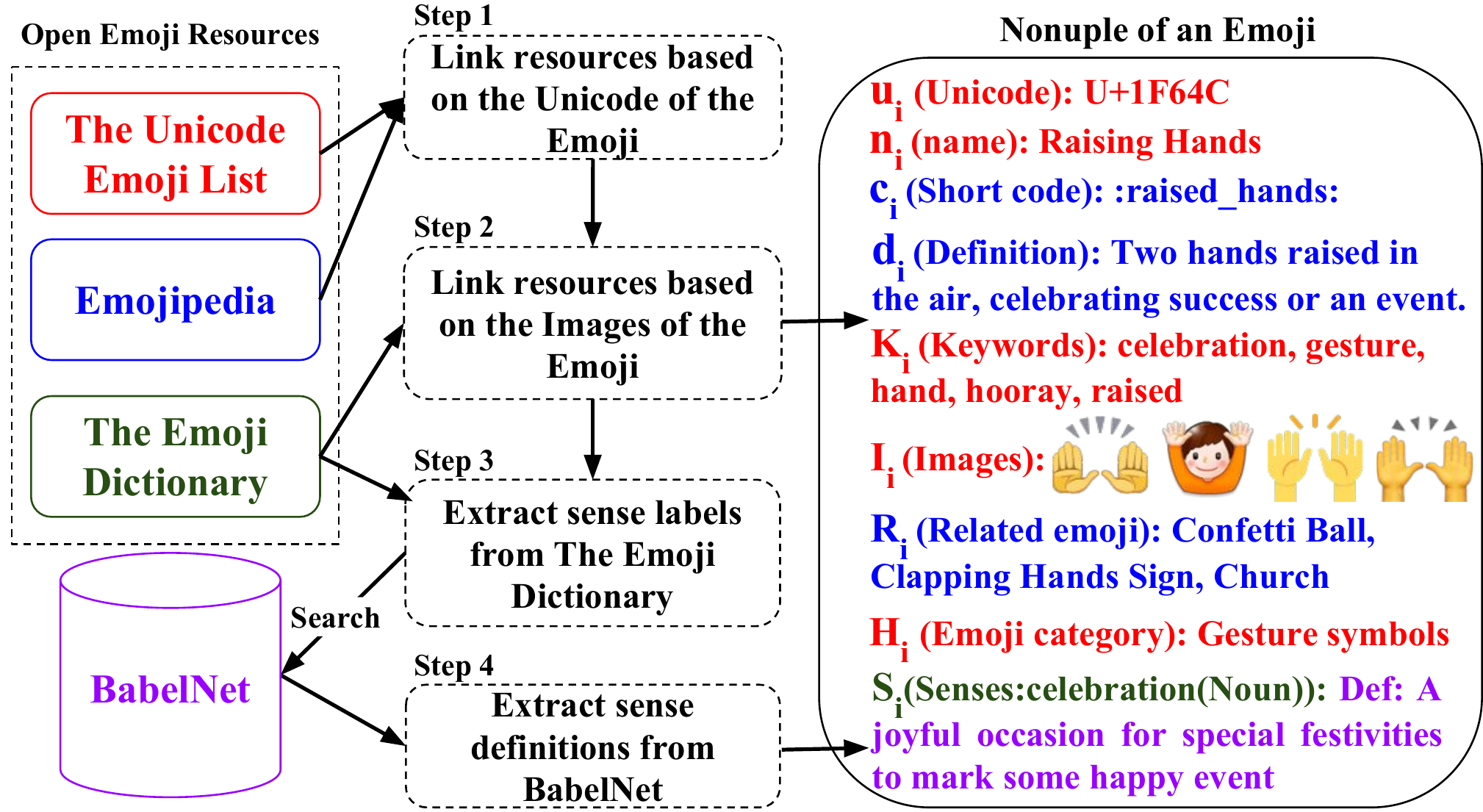} 
\caption{Construction of Emoji Representation in EmojiNet}
\label{buildingemojinet}
\end{figure}

An example of nonuple notation is shown as part of Figure~\ref{buildingemojinet}. Each element in the nonuple provides essential information on emoji and for emoji sense disambiguation. EmojiNet uses unicode $u_i$, name $n_i$, and short code name $c_i$ of an emoji $e_i \in E$ to uniquely identify $e_i$, and hence, to search EmojiNet. $d_i$ is a description of what is modeled in the emoji. It can sometimes help to understand the intended use of an emoji too. $K_i$ is also helpful to understand the intended uses of an emoji. $I_i$ helps to understand the rendering differences in each emoji based on different platforms. $R_i$ and $H_i$ could be useful to understand how emoji are related; thus, will be useful in tasks such as calculating emoji similarity and emoji sense disambiguation. Finally, $S_i$ holds all senses for $e_i$, including their POS tags and sense definitions and links them with BabelNet, which makes EmojiNet a machine-readable emoji sense inventory.

\subsection{Open resources used in EmojiNet}
A number of open resources, with appropriate permission from the dataset owners, are used to construct the nonuple of an emoji. This section introduces those resources and the information extracted from each of them. 

\subsubsection{The Unicode Consortium}
Unicode is an industry standard for the encoding, representation, and handling of text in computers which enables people around the world to use them in any language\footnote{\url{http://www.unicode.org/}}. The Unicode Consortium also maintains a complete list of the standardized Unicodes for each emoji\footnote{\url{https://goo.gl/lo3z1E}} along with other information on them such as manually curated keywords and images of emoji. Let the set of all emoji available in the Unicode emoji list be $E_U$. For each emoji $e_u \in E_U$, the Unicode character $u_u$ of $e_u$, the name $n_u$ of $e_u$, the set of all manually assigned keywords $K_{e_{u}}$ that describe the intended functionality of $e_u$, the set of all images $I_{e_{u}}$ associated with $e_u$ that are used to display $e_u$ on different platforms, and the set of categories $H_{e_{u}}$ which are all the categories that $e_u$ belongs to, are extracted from the Unicode Consortium website for inclusion in EmojiNet. 

\subsubsection{Emojipedia}
Emojipedia\footnote{\url{http://emojipedia.org/}} is a human-created emoji reference website that organizes emoji into a pre-defined sets of categories while also providing useful information about them. Specifically, for each emoji, Emojipedia lists the Unicode representation of the emoji, its short code, its variations over rendering platforms, and its relationships with other emoji. Let the set of all emoji available in Emojipedia be $E_{E}$. For each emoji $e_{e} \in E_{E}$, EmojiNet extracts the Unicode representation $u_{e}$, short code $c_{e}$, emoji definition $d_{e}$, and the set of related emoji $R_{e_{e}}$ of $e_{e}$ from Emojipedia.

\subsubsection{The Emoji Dictionary}
The Emoji Dictionary\footnote{\url{https://goo.gl/9gDVkE}} is the first crowdsourced emoji reference website that provides emoji definitions with their sense labels based on how they are used in sentences. It organizes the different meanings of an emoji under three part-of-speech tags, namely, nouns, verbs, and adjectives. It also lists an image of the emoji and its definition with example usages spanning across multiples senses with multiple part-of-speech tags. Let the set of all emoji available in The Emoji Dictionary be $E_{D}$. For each emoji $e_{d} \in E_{D}$, EmojiNet extracts its image $i_{e_{d}} \in I_{D}$, where $I_{D}$ is the set of all images of all emoji in $E_{D}$ and the set of crowd-generated sense labels $S_{e_{d}}$ from Emoji Dictionary.

\subsubsection{BabelNet}
BabelNet is the most comprehensive multilingual machine-readable semantic network available to date~\cite{navigli2010babelnet} and it has been shown useful in many research areas, including word sense disambiguation, semantic similarity, and sense clustering. It is a dictionary with a lexicographic and encyclopedic coverage of words within a semantic network that connects concepts in Wikipedia to the corresponding words in the BabelNet dictionary. Sense definitions from BabelNet are included in EmojiNet. For the set of all sense labels $S_{e_{d}}$ in each $e_{d} \in E_{D}$, EmojiNet extracts the sense definitions and examples for each sense label $s_{e_{d}} \in S_{e_{d}}$ from BabelNet.

\subsection{Resource Integration}
Figure~\ref{buildingemojinet} gives a high level four step overview of how the open resources are utilized to create an emoji representation. This section elaborates on each of these steps.

\subsubsection{Step 1 -- Linking resources based on the Unicode}
First, EmojiNet extracts all emoji characters that are currently supported by the Unicode Consortium and the information it stores for each one of them, such as emoji names, keywords and images. Then, for each emoji extracted from the Unicode website, EmojiNet extracts additional information such as the emoji short code, emoji description and related emoji from Emojipedia website. EmojiNet merges all information extracted from the two websites based on the Unicode representation of emoji and stores them under each emoji $e_i \in E$ using the nonuple notation described earlier.

\subsubsection{Step 2 -- Linking resources based on the Images}
The Emoji Dictionary does not store the Unicode character representations of emoji, hence integrating it with the emoji data extracted from the Unicode Consortium and Emojipedia websites is done based on matching the images of the emoji available in the three resources. For this purpose, we extracted 18,615 images representing all 2,389 emoji from the Unicode Consortium website and created an index, which we refer to as our example set, $I_{x}$. We also downloaded images of all emoji listed on The Emoji Dictionary website, which resulted in a total of 1,074 images, from which we created our test image dataset, $I_{t}$. We implemented a nearest neighborhood-based image matching algorithm based on~\cite{santos2010java} that matches each image in $I_{t}$ with the images in $I_{x}$. This algorithm has shown to perform well when aligning images with few colors and objects, which is the case with emoji. Since images are of different resolutions, we first normalized them into a 300x300px space and then divided them along a lattice of 25 non-overlapping regions of size 25x25px. We then calculated the average color intensity of each region by averaging its $R$, $G$ and $B$ pixel color values. To calculate the dissimilarity between two images, we summed the $L_2$ distance of the average color intensities of the corresponding regions. We selected $L_2$ distance as it prefers many medium disagreements to one large disagreement as in $L_1$ distance. The final accumulated value that we received for a pair of images will be a measure of the dissimilarity of the two images. For each image in $I_t$, the least dissimilar image from $I_x$ is chosen and the corresponding emoji nonuple information is merged.

\subsubsection{Step 3 -- Extracting sense labels}
\begin{figure}
\centering
\includegraphics[width=1.0\linewidth]{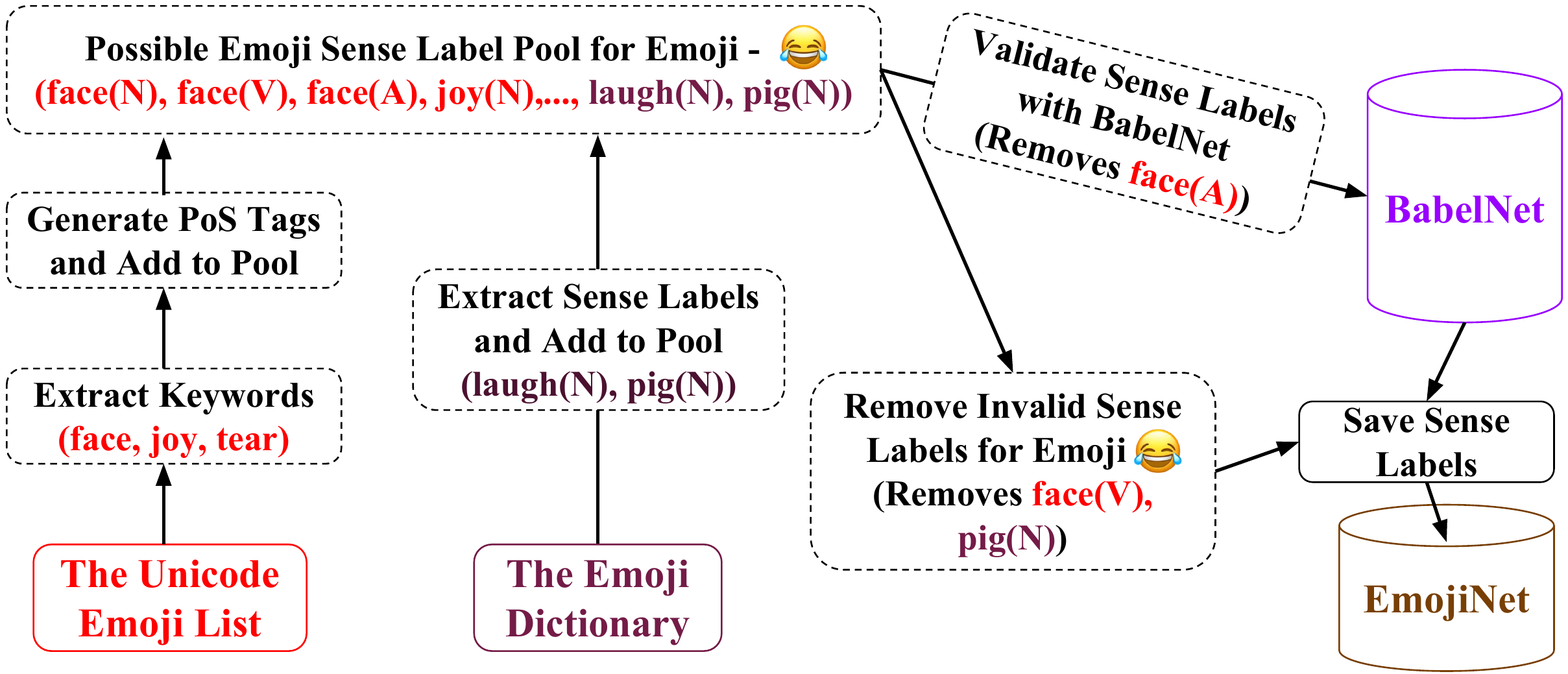} 
\caption{Using The Unicode Consortium and The Emoji Dictionary for Sense Label Filtering}
\label{sensefiltering}
\end{figure}

The Emoji Dictionary lists sense labels for each emoji which were obtained through its crowdsourced data collection platform, while the Unicode Consortium lists intended meanings for each emoji as keywords, but without any part-of-speech tags. The two resources thus carry complementary information about emoji meanings necessary to create a sense label. EmojiNet follows the procedure illustrated in Figure~\ref{sensefiltering} to extract emoji sense labels using the two resources. For each emoji, EmojiNet extracts all the emoji sense labels listed in The Emoji Dictionary.  This resulted in a total of 31,944 sense labels. For each of the 6,057 keywords for emoji listed in the Unicode Consortium website, it then generates three sense labels using the three part-of-speech tags; noun, verb, and adjective. That means, for a keyword listed in the Unicode Consortium website such as \texttt{face}, EmojiNet generates the three sense labels \texttt{face(N)}, \texttt{face(V)}, and \texttt{face(A)} as shown in Figure~\ref{sensefiltering}. We selected only three part-of-speech tags as they were the only part-of-speech tags supported by The Emoji Dictionary and other emoji sense inventories~\cite{emojinet}.

Following the above step, a total of 18,171 sense labels for the 6,057 keywords are created. Next, EmojiNet combines the sense labels from the two resources into a pool of sense labels, totaling 50,115 sense labels in the pool. However, not all senses in the pool of sense labels are valid. For example, the sense label \texttt{face(A)} is invalid as the word \texttt{face} cannot be used as an adjective in the English language. To filter out invalid sense labels from the sense label pool, EmojiNet validates each sense label in the pool against the valid sense labels in BabelNet sense inventory. During this validation process, a total of 21,779 sense labels were discarded from the sense label pool where 10,848 of them were extracted from The Emoji Dictionary and 10,931 of them were generated from the Unicode Consortium keywords. The above filtering step leaves 28,336 valid sense labels in the sense label pool. We also noticed that there are a lot of sense labels in the pool that do not represent valid meanings for certain emoji. For example, for the emoji \emoji{1F602}, \texttt{pig(N)}, \texttt{rainbow(N)}, and \texttt{face(V)} are listed among many other invalid meanings. Even though these are valid English sense labels, they are not valid meanings for the \emoji{1F602} emoji, thus we remove such instances. Most of such invalid sense labels were extracted from The Emoji Dictionary, and due to non availability of input validation methods in The Emoji Dictionary website, those being ended up adding to Emoji Dictionary's sense inventory. With the help of two human annotators, we were able to remove a total of 15,432 such sense labels. The remaining 12,904 sense labels are ready to be assigned their sense definitions using BabelNet, as described next.

\subsubsection{Step 4 -- Extracting and linking with BabelNet senses}
For a given sense label, there could be multiple sense definitions available in BabelNet. For example, the current version of BabelNet lists 6 different sense definitions for the sense label \texttt{laugh(noun)}. Thus, to select the most appropriate sense definition out of the multiple BabelNet sense definitions and link them with the sense labels extracted in Step 3, a Word Sense Disambiguation (WSD) task needs to be performed. To conduct this WSD task, we use the Manually Annotated Sub-Corpus (MASC)\footnote{\url{https://goo.gl/OeLc2F}} with a most frequent sense (MFS) baseline. We choose the MASC corpus because it is a balanced dataset that represents different text categories such as tweets, blogs, emails, etc. and Moro {\em{et al.}} have already annotated it using the BabelNet senses~\cite{moro2014annotating}. A MFS baseline outputs the MFS calculated for each word with respect to a sense-annotated text corpus. Then the baseline assigns the MFS of a word as its correct word sense. We use a MFS-based WSD baseline due to the fact that MFS is a very strong, hard-to-beat baseline model for WSD tasks~\cite{basile2014enhanced}.

\begin{figure}
\centering
\includegraphics[width=1.0\linewidth]{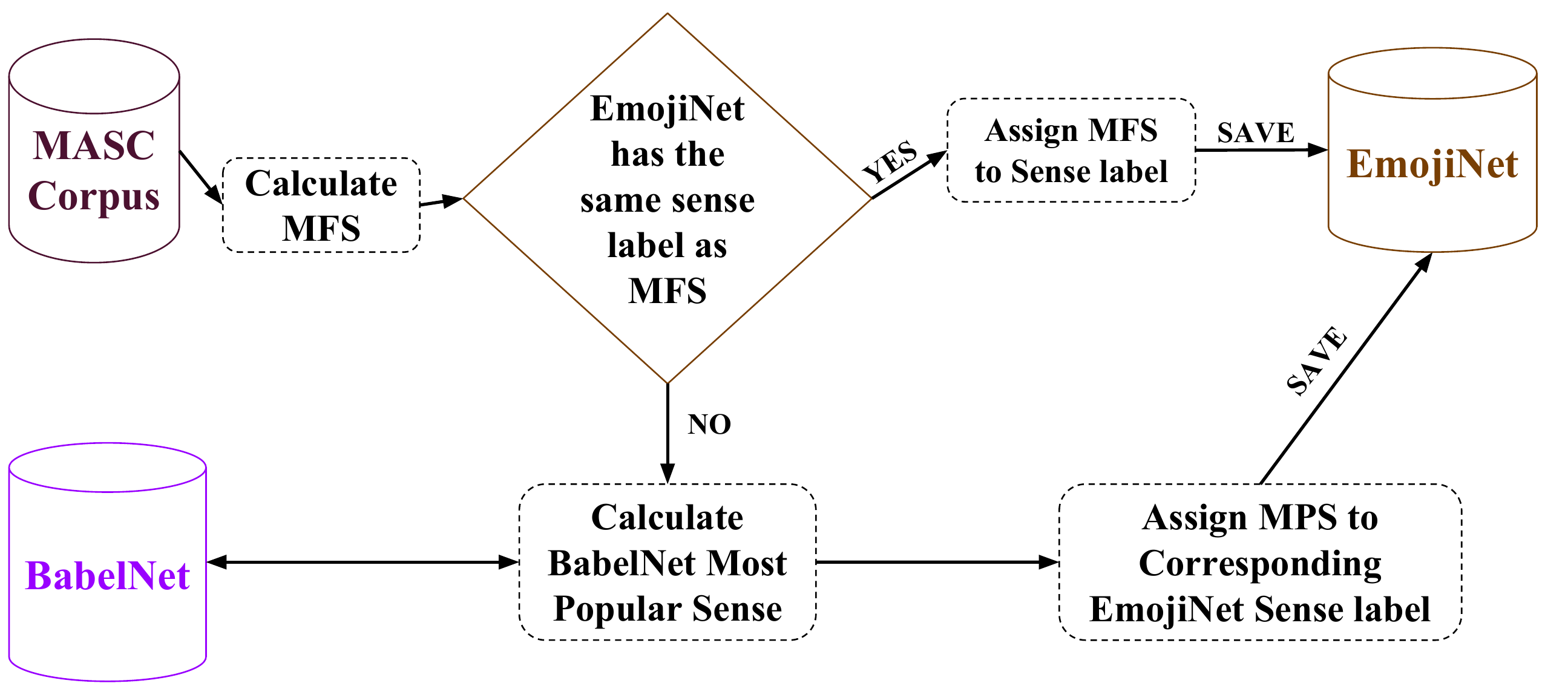} 
\caption{Assigning BabelNet Sense Definitions}
\label{wsd}
\end{figure}

Figure~\ref{wsd} depicts the process of assigning BabelNet sense definitions to the sense labels in EmojiNet. We use the MASC annotations provided by Moro {\em{et al.}} to calculate the MFS of each word in the MASC corpus. Then, for all sense labels that are common for the MASC corpus and EmojiNet, we assign the calculated MFS for each of them as their corresponding sense definition and save their sense definitions in EmojiNet. However, not all sense labels in EmojiNet were assigned BabelNet senses in the above WSD task as several sense labels were not present in the MASC corpus. To assign sense definitions for those that were missed in the earlier WSD task, we defined a second WSD task based on the most popular sense (MPS) of each BabelNet sense. We define the MPS of a sense label as follows. For each BabelNet sense label $B_s$, we take the count of all sense definitions BabelNet lists for $B_s$. The MPS of $B_s$ is the BabelNet sense ID that has the highest number of definitions for $B_s$. For sense labels that there are more than one MPS available in BabelNet, we manually assign the correct BabelNet sense ID. Once the MPS is calculated, those will be assigned to their corresponding sense labels in EmojiNet which were left out in the MFS-based WSD task.

\section{Enhancing EmojiNet for Analysis Tasks} \label{sec:extend}

Making EmojiNet more useful beyond just serving as a machine-readable sense inventory and to enable its use for emoji analysis tasks on ill-formatted social media text required the following enhancements.

\subsubsection{Adding word embeddings to EmojiNet}

As pointed out earlier, the accuracy of the sense disambiguation tasks can be improved by incorporating more context words~\cite{vasilescu2004evaluating}, and NLP tools trained on well-formed text might not work well with the language variations seen in social media~\cite{ritter2011named}. Thus, to make EmojiNet a more robust tool for working with social media text processing, we derive additional context words based on word embedding models learned over Twitter and news articles, respectively.

We collected a Twitter dataset that contained emoji using the Twitter streaming API\footnote{\url{https://dev.twitter.com/streaming/public}} to train a word embedding model on tweets with emoji. The dataset was collected using emoji Unicodes as filtering words, over a four week period, starting from 6$^{th}$ August 2016 to 8$^{th}$ September 2016. It consists of 147 million tweets containing emoji. We removed all retweets and used the remaining 110 million unique tweets for training purposes. When training the Twitter-based word embedding model~\cite{DBLP:journals/corr/abs-1301-3781}, we first convert all emoji into textual features using Emoji for Python\footnote{\url{https://pypi.python.org/pypi/emoji/}} API. Then we remove all stopwords and perform stemming across all tweets. We then feed all the training data (i.e. words found in tweets, including emoji) to the Word2Vec tool and train it using a Skip-gram model with negative sampling. We choose Skip-gram model with negative sampling to train our model as it is shown to generate robust word embedding models even when certain words are less frequent in the training corpus~\cite{NIPS2013_5021}. We set the number of dimensions of our model to 300 and the negative sampling rate to 10 sample words, which works well with medium-sized datasets~\cite{NIPS2013_5021}. We set the context word window to be 5 so that it will consider 5 words to left and right of the target word at each iteration of the training process. This setting is suitable for sentences where average sentence length is less than 11 words, as is the case in tweets~\cite{HuTK13}. We ignore the words that occur fewer than 10 times in our Twitter dataset when training the word embedding model. We use a publicly available word embedding model that is trained over Google News corpus\footnote{\url{https://goo.gl/QaxjVC}} to learn additional context words for emoji sense definitions.

\begin{table}[]
\centering
\caption{EmojiNet Statistics}
\label{emojinetstats}
\begin{tabular}{|l|c|c|}
\hline

\multicolumn{1}{|c|}{\textbf{Emoji feature}} & \textbf{\begin{tabular}[c]{@{}c@{}}\# of emoji \\ with each feature\end{tabular}} & \textbf{\begin{tabular}[c]{@{}c@{}}\# of data stored \\ for each feature\end{tabular}} \\ \hline
Unicodes                                       & 2,389                                                                             & 2,389                                                                                  \\ \hline
Emoji Names                                    & 2,389                                                                             & 2,389                                                                                  \\ \hline
Short Codes                                    & 2,389                                                                             & 2,389                                                                                  \\ \hline
Descriptions                                   & 2,389                                                                             & 2,389                                                                                  \\ \hline
Keywords                                       & 2,389                                                                             & 6,057                                                                                  \\ \hline
Images                                         & 2,389                                                                             & 18,615                                                                                 \\ \hline
Related Emoji                                  & 1,755                                                                              & 7,544                                                                                  \\ \hline
Categories                                     & 2,389                                                                             & 141                                                                                  \\ \hline
Senses                                         & 2,389                                                                             & 12,904                                                                                 \\ \hline
\end{tabular}
\end{table}

We follow an approach similar to the one presented by Eisner {\em{et al.}} when learning additional context words for emoji sense definitions~\cite{eisner2016emoji2vec}. For each emoji $e_i \in E$, we extract the definition $d_i$ of the emoji $e_i$ and the set of all emoji sense definitions $S_i$ of $e_i$ from EmojiNet. Then, for each word $w$ in $d_i$, we extract the twenty most similar words from the two word embedding models as two separate sets, namely $CW_{e_i}^T$ and $CW_{e_i}^N$. For example, for \emoji{1F52B}, EmojiNet lists {\em{``A gun emoji, more precisely a pistol. A weapon that has potential to cause great harm''}} as its emoji definition. To generate context words, we replace each word in the definition above with the top twenty most similar words learned for it using the two word embedding models, respectively. We do the same for each emoji sense definition for \emoji{1F52B} as well. For each emoji sense definition $s_i \in S_i$ that belongs to $e_i$, we then extract the words $w_{s_i}$ in $s_i \in S_i$ and repeat the same process to learn two separate context word sets $CW_{e_i-s_i}^T$ and $CW_{e_i-s_i}^N$, based on the twenty most similar words for each word $w_{s_i}$ in $s_i \in S_i$. The separate sets allow us to mark if a context word was learned from social media (Twitter) or more well-formed text (news articles) in EmojiNet.

\subsubsection{Adding platform-specific meanings to EmojiNet}

As pointed out by~\cite{miller2016blissfully}, platform-specific emoji meanings could also play an important role in emoji understanding tasks. We came up with a list\footnote{\url{https://goo.gl/bir9xV}} of 40 most confused emoji based on the differences in their platform-specific images and crowd-provided senses, including the 25 emoji studied by Miller {\em{et al.}} We setup an Amazon Mechanical Turk (AMT) experiment to identify the platform-specific meanings associated with the 40 selected emoji. We selected five vendor platforms for our study, namely Google, Apple, Windows, Samsung, and Twitter, and extracted all of the emoji sense labels stored in EmojiNet. In our experiment, a single AMT task asks a worker to say whether they think that a given sense label is a reasonable sense for a platform-specific emoji. Radio buttons (agree or disagree) are used to record their decisions, along with a text field to give a brief explanation. Results with no, repeating, or nonsense explanations were filtered away under the assumption that the worker was a spammer. We conducted a total of 14,448 such AMT tasks, of which 1,128 were filtered as spam.

We looked at what were the emoji that had platform specific meanings. We specifically look for meanings that are unique for certain emoji when they were shown in certain platforms. We found 27 emoji (67.50\%) that had platform specific meanings. For example, for \emoji{1F60D}, we noticed that windows platform was the only one that shows a smiling face with teeth displaying as shown in \emoji{26D1}. Therefore, the AMT workers have assigned laugh as a meaning for \emoji{26D1} but not for any other emoji for the same Unicode representation including \emoji{1F60D}. We also noticed that the Samsung platform-based emoji had the least number of meanings associated with their images, which tells us Amazon workers had hard time agreeing with each other on the emoji meanings when Samsung images are displayed. Google platform images had the highest agreement among emoji senses. We've added these vendor-specific meanings into EmojiNet dataset.

\begin{figure}
\centering
\includegraphics[width=1.0\linewidth]{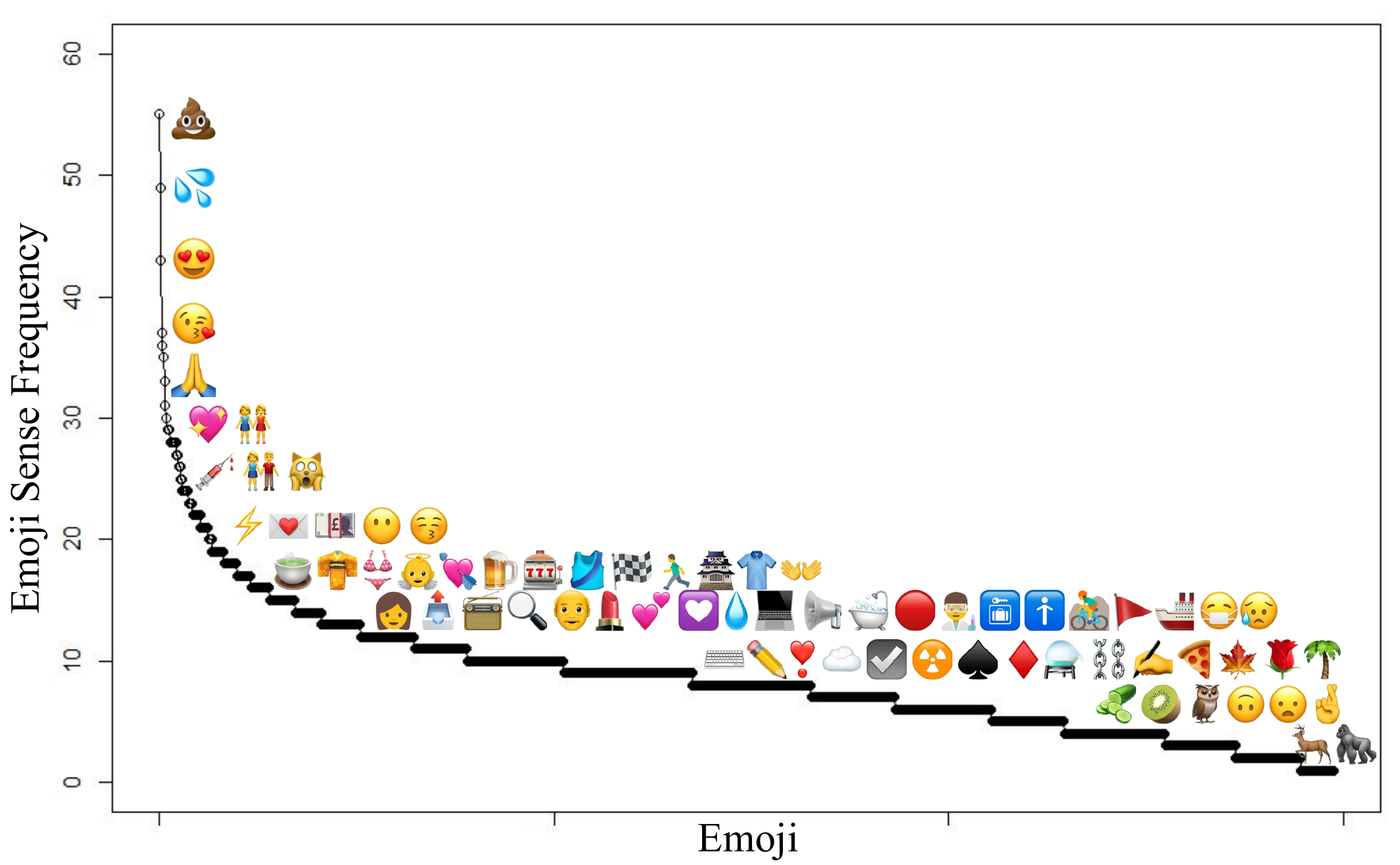} 
\caption{Emoji Sense Distribution}
\label{graph}
\end{figure}

\subsubsection{EmojiNet Web application and REST API}

To make it easy to browse and programmatically access the EmojiNet dataset, we host EmojiNet as a web application at \url{http://emojinet.knoesis.org/}. The web application supports searching the EmojiNet dataset based on the Unicode character representation, name, or the short code of an emoji. It also supports browsing EmojiNet by specifying the part-of-speech tags of the emoji senses. A REST API is implemented so that computer applications can also access the EmojiNet dataset. The API has a series of methods that can be invoked over an HTTP connection that return data in a JSON object format. The resource, along with the complementary sense embedding, vendor-specific sense data and REST API with documentation, is freely available to the public for research use. 

Table~\ref{emojinetstats} lists some summary statistics for the data stored in EmojiNet and the emoji data distribution. Each emoji in EmojiNet carries all features listed in Table~\ref{emojinetstats} except related emoji. A total of 7,544 related emoji pairs have been stored in EmojiNet that belongs to 1,755 emoji. There are 6,057 emoji keywords, 18,615 images, 141 categories and 12,904 sense labels being shared across the 2,389 emoji. An emoji in EmojiNet has 5 to 6 senses on average.  

Figure~\ref{graph} plots the number of unique emoji senses for each emoji stored in EmojiNet. A selected set of emoji are also shown there. Emoji with a diverse set of senses include \emoji{1F4A9} (55) followed by \emoji{1F4A6} (49). For example, \emoji{1F4A9} had senses ranging from chocolate to smelly. When looking at the senses, it was evident that most of the senses are based on the look and feel of the emoji. For example, \emoji{1F4A9} had many sense variations that interpret as feces, and some sense variations which were based on the color and the shape of the emoji. \emoji{1F4A6} had senses ranging from sweat to rain. We also noticed \emoji{1F60D}, \emoji{1F618} and \emoji{1F602}, which are three of the most popular emoji on Twitter\footnote{\url{http://www.emojitracker.com/}}, were among EmojiNet's top 10 emoji with most number of senses. We examined the emoji that had least number of emoji senses (1). Those include blood type emoji such as \emoji{1F170} and \emoji{1F18E}, buttons such as \emoji{1F192} and \emoji{1F193}, and newly introduced emoji such as \emoji{1F98C} and \emoji{1F98D}. We found that all of them do not exist in The Emoji Dictionary website, hence they did not have any crowd-generated emoji meanings saved in EmojiNet. We also noticed that they have only one keyword listed in the Unicode Consortium website as the intended meaning. Some of them, such as animal faces, were recently introduced.

\section{Dataset Evaluation} \label{sec:eval}

This section evaluates the process we used to curate the data published in EmojiNet. In particular, we evaluate the nearest neighborhood-based image processing algorithm that we used to integrate emoji resources and the most frequent sense-based and most popular sense-based WSD algorithms that we used to assign meanings to emoji sense labels.

\subsection{Resource Integration Evaluation}
We evaluate how the nearest neighborhood-based image processing algorithm matches images from The Emoji Dictionary website (i.e. $I_{t}$) with the images downloaded from the Unicode Consortium website (i.e. $I_{x}$). The Unicode Consortium website contains multiple vendor-specific images for a given emoji that depict how an emoji looks on those vendors' platforms (i.e. different emoji for different platforms such as Apple, Samsung, Windows, Twitter etc.). Since we have indexed all vendor-specific images of each emoji under that emoji's Unicode representation, to correctly map an image in $I_{t}$ with $I_{x}$, we only require an image in $I_{t}$ to match with any one of those vendor-specific images in $I_{x}$ for a given emoji. Once the matching process is completed, we pick the top ranked match for each emoji based on the dissimilarity of the two
matched images and manually evaluate them for equality. The Unicode representation of the top ranked matched image from $I_{x}$ will be assigned as the Unicode representation of the matching image from $I_{t}$. Though the image processing algorithm we used is naive, it works well for our study as the images of the emoji are not complex (i.e. each image has one object such as a face or a fruit) and they do not contain complex color combinations (i.e. one or two colors with a transparent/white background). The image processing algorithm we used combines color (spectral) information with spatial (position/distribution) information and tends to represent those features well when the images are simple.

Out of the 1,074 image instances we checked, our algorithm could correctly find matching images for 1,034 images in $I_{t}$ with an accuracy of 96.27\%. An error analysis performed on the 40 incorrect matches revealed that 13 family emoji, 9 person emoji and 8 clock emoji were identified incorrectly among others. These image types had minimal differences in either objects or color, hence the algorithm had failed to match them correctly. For example, 9 person emoji had very slight differences in objects such as long hair in one image versus the short hair in the other. The clock emoji had their arms at different locations while the color of the images were identical. We manually corrected the 40 incorrect matches and assigned them their correct Unicode representations to complete the integration between The Emoji Dictionary data with Unicode Consortium data.

\begin{table}[]
\centering
\caption{Word Sense Disambiguation Statistics}
\label{wsdstats}
\begin{tabular}{|l|c|c|c|}
\hline
                                   & \textbf{Correct}         & \textbf{Incorrect}     & \textbf{Total} \\ \hline
Noun                               & 6,633 (86.64\%)            & 1,022 (13.36\%)            & 7,655          \\ \hline
Verb                               & 2,231 (77.14\%)            & 661 (22.86\%)           & 2,892            \\ \hline
Adjective                          & 1,915 (81.24\%)            & 442 (18.76\%)            & 2,357            \\ \hline
\multicolumn{1}{|l|}{\textbf{Total}} & \textbf{10,779 (83.53\%)} & \textbf{2,125 (16.47\%)} & \textbf{12,904} \\ \hline
\end{tabular}
\end{table}

\begin{table*}[]
\centering
\caption{Top 10 Emoji based on the Emoji Sense Disambiguation Accuracy (in \% values)}
\label{sensedisstats}
\begin{tabular}{|l|c|c|c|c|c|c|c|c|c|c|c|}
\hline
\multicolumn{1}{|c|}{\textbf{Emoji}} & \emoji{1F609} & \emoji{1F606} & \emoji{1F605} & \emoji{1F64F} & \emoji{1F604} & \emoji{1F634} & \emoji{1F602} & \emoji{1F633} & \emoji{1F64C} & \emoji{1F601} & \textbf{Avg. Accuracy} \\ \hline
\textbf{BabelNet-based}                       & 24.48 & 20.93 & 16.27 & 12.00 & 16.66 & 18.75 & 15.21 & 20.45 & 12.00 & 27.08 & 18.38 \\ \hline
\textbf{Twitter-based}                       & \textbf{61.22} & \textbf{60.00} & \textbf{56.41} & \textbf{56.00} & 43.58 & \textbf{51.21} & \textbf{48.57} & \textbf{47.72} & \textbf{46.51} & \textbf{44.18} & \textbf{51.54} \\ \hline
\textbf{News-based}                       & 32.65 & 59.45 & 41.46 & 29.16 & \textbf{52.17} & 41.17 & 43.24 & 13.63 & 37.50 & 38.63  & 38.91 \\ \hline
\end{tabular}
\end{table*}

\subsection{Sense Assignment Evaluation}

Here we discuss how we evaluated the MFS-based and MPS-based WSD algorithms that we used to link Emoji sense labels with BabelNet sense definitions. To do this evaluation, we sought the help of two human judges who have experience in NLP research. We provided them with all emoji included in EmojiNet, listing all the valid sense labels for each emoji and their corresponding BabelNet senses (BabelNet sense IDs with definitions) calculated though either MFS or MPS baselines. They were asked to mark whether they thought that the suggested BabelNet sense ID was the correct sense ID for the emoji sense label listed. We calculated the agreement between the two judges for this task using Cohen's kappa coefficient\footnote{\url{https://goo.gl/szv50P}} and obtained an agreement value of 0.7134, which is considered to be a good agreement.

Out of the 12,904 sense labels we provided them to disambiguate, 7,815 appeared in both the EmojiNet dataset and MASC dataset, so they were assigned BabelNet sense definitions through the MFS-based WSD approach. Our judges evaluated the sense assignments based on whether they thought that the suggested BabelNet sense ID assigned by the MFS baseline was the correct sense ID for the emoji sense label. They decided that 6,673 sense labels were assigned correct BabelNet sense IDs, yielding an accuracy of 85.38\% for the MFS baseline. Judges then assigned the correct sense IDs for the 1,142 sense labels that were sense disambiguated incorrectly by the MFS baseline. The remaining 5,089 sense labels which were not assigned senses by the MFS baseline were considered for a second WSD task based on a MPS baseline. While evaluating the accuracy of the MPS baseline, the judges followed the same approach that they followed for evaluating the MFS baseline. Based on the evaluation results, we found that the MPS baseline has achieved 80.68\% accuracy in the WSD task. There were 983 sense labels which were sense disambiguated incorrectly in this approach, which were then corrected by the judges. Overall, the two WSD baselines have correctly sense disambiguated a total of 10,779 sense labels, yielding an accuracy of 83.53\% for the WSD task. Table~\ref{wsdstats} integrates the results obtained by both word sense disambiguation algorithms for different part-of-speech tags. The results shows that the two WSD approaches we used have performed reasonably well in disambiguating the sense labels in EmojiNet.

\section{Applications} \label{sec:usecase}
In this section, we illustrate two applications of EmojiNet, namely, emoji sense disambiguation and emoji sense similarity evaluation. Note that the intention of this section is to demonstrate EmojiNet's utility to researchers, rather than to propose comprehensive solutions to these challenging tasks.

\subsection{Emoji Sense Disambiguation}
Emoji sense disambiguation is the ability to identify the meaning of an emoji in the context of a message in a computational manner. Previous research has identified the importance of the problem~\cite{emojinet}, however have not solved it. To solve the emoji sense disambiguation problem, there has to be an emoji sense inventory that a computer program could use to extract emoji meanings.

To show EmojiNet could be used as a sense inventory for emoji sense disambiguation, 
we first select 25 emoji which have shown to be interpreted differently when used in communication by previous research~\cite{miller2016blissfully}. Then we take the Twitter corpus that we used to train the word embedding model discussed in Section~\ref{sec:extend} and randomly select 50 tweets for each of the 25 emoji. We select tweets that contain only one emoji anywhere in the middle of the tweet. To disambiguate the sense of an emoji in a tweet, we compare the context of the emoji in the tweet with the contexts of each emoji sense for that emoji obtained from EmojiNet. This tweet context is defined as all words surrounding the emoji. 
We define three sets of contexts for an emoji sense based on the three different datasets we used to generate them: 
(i) {\em BabelNet-based context}: This is the set of words coming from BabelNet sense definitions which we extracted for an emoji, (ii) {\em Twitter-based context}: This is the set of context words learned by using the Twitter word embedding model for the emoji from Section~\ref{sec:extend}, and (iii) {\em News-based context}: This is the set of context words learned by using the Google News word embedding model for an emoji from Section~\ref{sec:extend}. To find the sense of an emoji in a tweet, we calculate the context overlap between the context of the emoji in the tweet with the context words taken from each of the above three sets. Following past studies, the sense with the highest context word overlap is assigned to the emoji at the end of a successful run of the algorithm~\cite{vasilescu2004evaluating}. We then asked two human judges to evaluate the emoji senses assigned to the emoji in our tweet dataset. We asked the judges to label the sense assignment as {\em correct} if they think that the chosen sense for an emoji in a tweet is the most appropriate sense that could be assigned to it from EmojiNet or {\em incorrect} if they do not think the sense is appropriately assigned for the emoji in a tweet. The agreement between the two judges for this task measured by Cohen's kappa coefficient was 0.6878, which is considered to be a good agreement.

Table~\ref{sensedisstats} lists the top 10 emoji based on their sense disambiguation accuracy. We define the emoji sense disambiguation accuracy for an emoji as the ratio between the number of correctly sense disambiguated messages (tweets) and the number of total sense disambiguated messages for that emoji. Among the 25 emoji in our dataset, \emoji{1F609} gives the highest sense disambiguation accuracy of 0.61. We observe that Twitter-based context vectors outperforms the other two context vectors constantly, except for disambiguating the sense of \emoji{1F601}. This observation aligns with what past research on social media text processing suggest us, which is, tools designed for well-formed text processing will not work well when used for ill-formatted text processing~\cite{ritter2011named}. The average number of Twitter-based context words for an emoji sense definition was very high compared to that of BabelNet-based contexts. This align with the past research too, that the disambiguation results can be improved when we increase the number of context words in the sense definitions~\cite{vasilescu2004evaluating}. 
These evaluation results validates the importance of the improvements we made to EmojiNet by introducing context word vectors learned by Twitter and Google News corpuses.

\begin{figure}
\centering
\includegraphics[width=1.0\linewidth]{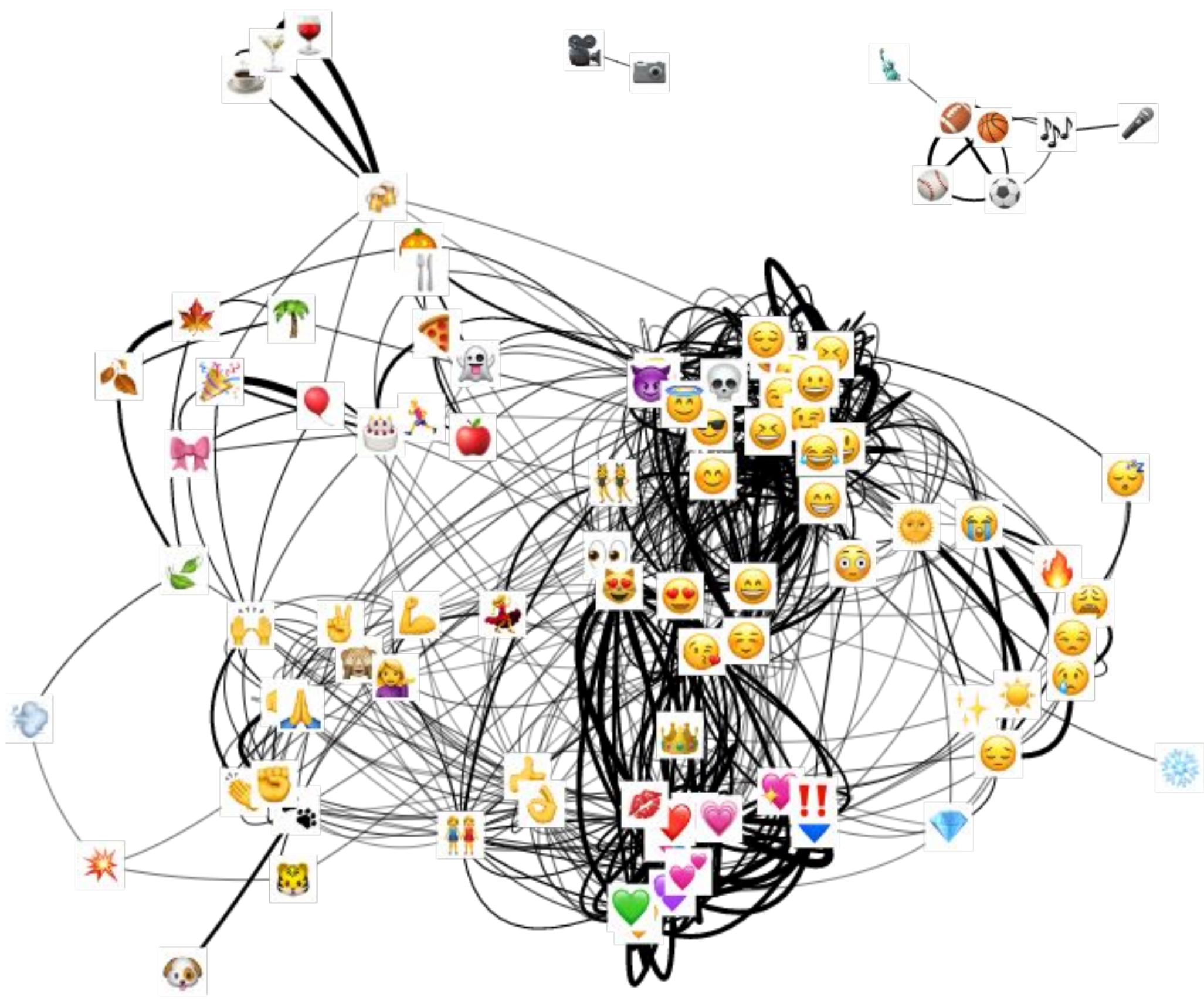} 
\caption{Emoji Clusters using Emoji Sense Overlap}
\label{emojisim}
\end{figure}

\subsection{Emoji Sense Similarity}

Similar to semantic similarity of words\footnote{\url{https://goo.gl/ITXkAT}}, we define emoji similarity based on the likeness of their meaning as defined by the sense labels assigned to each emoji. This is a new notion of `emoji similarity' compared to previous work that defined similarity by emoji functionality or topic~\cite{barbieri2016does,eisner2016emoji2vec} and is uniquely enabled by EmojiNet's sense repository. 
This sense similarity is similar to how semantic similarity measures have been defined using sense inventories such as WordNet\footnote{\url{https://wordnet.princeton.edu/}} and BabelNet. 
Since EmojiNet carries functional and topical emoji meanings available in the Unicode Consortium and The Emoji Dictionary websites in addition to the other intended meanings of emoji if any, our method  complements other similarity measures. We next describe a use-case where we model an emoji similarity graph using the emoji present in EmoTwi50 dataset created by Barbieri {\em{et al.}} to explain sense-based emoji similarity.

EmoTwi50 is a dataset that contains 25 manually created and 25 randomly created emoji pairs, totaling 100 unique emoji. Barbieri {\em{et al.}} created and used this dataset to find functional and topical similarity of the 50 emoji pairs. We use it to create our emoji similarity graph based on emoji senses. We first extract the sense labels of the 100 emoji in EmoTwi50 dataset from EmojiNet. A node in the emoji similarity graph is an emoji and an edge exists if there is at least one common sense between them. Figure~\ref{emojisim} visualizes this graph, with the thickness of an edge corresponding to the number of shared emoji senses between them. We then run a label propagation community detection algorithm~\cite{barber2009detecting} to identify emoji communities (clusters) based on their sense overlap. This revealed 16 clusters in our graph, each of which represents `sense-similar' emoji. Due to space limitations, we list a selected set of emoji within different clusters and label them in Table~\ref{emojiclusters}. We can see that the smiling face emoji have been clustered together while sad faces, hearts, drinks and hand symbols form their own clusters. We also notice two islands, which we have labeled as cameras and sports \& entertainment.

Once the graph is computed, we can use any traditional semantic, path or set similarity measure to find the sense similarity between any two emoji in the graph. For example, we use Jaccard Similarity\footnote{\url{https://goo.gl/RKkRzF}} to find the sense similarity between \emoji{1F603} and \emoji{1F600}. Both emoji have 12 sense labels each, shares 9 sense labels, and have 15 unique sense labels between them. Thus, the sense similarity between them can be calculated as the ratio between 9 and 15, which gives 0.60. Table~\ref{emojisimtable} lists the ten most similar emoji pairs calculated based on EmoTwi50 dataset. We can replace Jaccard Similarity with a sophisticated similarity measure to improve the results shown. The emoji similarity dataset we created using Jaccard Similarity is available to download at \url{http://emojinet.knoesis.org/}.

\begin{table}[]
\centering
\caption{Selected Emoji Sense Clusters in EmoTwi50}
\label{emojiclusters}
\begin{tabular}{|l|l|}
\hline
\multicolumn{1}{|c|}{\textbf{Cluster Name}} & \multicolumn{1}{c|}{\textbf{Emoji}} \\ \hline
Smiling Faces                                & \emoji{1F600} \emoji{1F601} \emoji{1F602} \emoji{1F603} \emoji{1F604} \emoji{1F605} \emoji{1F606}                               \\ \hline
Love                                & \emoji{2764} \emoji{1F495} \emoji{1F497} \emoji{1F498} \emoji{1F49E} \emoji{1F499} \emoji{1F49A}                               \\ \hline
Sad Faces                                & \emoji{1F62D} \emoji{1F622} \emoji{1F629} \emoji{1F612} \emoji{1F614} \emoji{1F634}                                \\ \hline
Drinks                                & \emoji{1F377} \emoji{1F378} \emoji{1F37B} \emoji{2615}                               \\ \hline
Cameras                                & \emoji{1F3A5} \emoji{1F4F7}                               \\ \hline
Sports \& Entertainment                                & \emoji{26BE} \emoji{1F3C0} \emoji{1F3C8} \emoji{26BD} \emoji{1F3A4} \emoji{1F3B5} \emoji{1F5FD}                               \\ \hline

\end{tabular}
\end{table}

\begin{table*}[]
\centering
\caption{Ten Most Similar Emoji Pairs Based on Jaccard Similarity}
\label{emojisimtable}
\begin{tabular}{ccccccccccc}
{\textbf{Emoji Pair}} & \emoji{1F603} \emoji{1F600} & \emoji{1F49E} \emoji{1F499} & \emoji{1F49E} \emoji{1F495} & \emoji{1F49A} \emoji{1F495} & \emoji{1F49E} \emoji{1F49A} & \emoji{1F62D} \emoji{1F622} & \emoji{1F389} \emoji{1F388} & \emoji{1F604} \emoji{1F603} & \emoji{1F61D} \emoji{1F61C} & \emoji{1F606} \emoji{1F603}  \\

{\textbf{Similarity}} & 0.60 & 0.57 & 0.56 & 0.52 & 0.52 & 0.50 & 0.50 & 0.50 & 0.48 & 0.47 %

\end{tabular}
\end{table*}

\section{Conclusion and Future Work} \label{sec:con}

This paper presented the release of EmojiNet, the largest machine-readable emoji sense inventory that links Unicode emoji representations to their English meanings extracted from the Web. We described how (i) three open resources were integrated to build EmojiNet, (ii) word embedding models and vendor-specific emoji senses were used to improve EmojiNet, and (iii) how the resource building process was evaluated. We developed a web application to browse the EmojiNet dataset, a REST API to access the functionality of EmojiNet, and showed how EmojiNet can be used to solve emoji sense disambiguation and emoji sense similarity problems. In the future, we plan to extend the sense inventory by adding machine processable emoji meanings that were not present in BabelNet but listed as intended meanings by the Unicode Consortium. We also want to introduce a semi-automatic update process to keep EmojiNet up-to-date as and when new emoji are supported by the Unicode Consortium. We would also like to work on introducing more sophisticated algorithms to solve emoji sense disambiguation and emoji sense similarity problems in the future to support better understanding of emoji use over social media.

\section{Acknowledgments}
We are grateful to Nicole Selken, the designer of The Emoji Dictionary and Jeremy Burge, the founder of Emojipedia for giving us the permission to use their web resources for our research. We are thankful to Scott Duberstein for helping us with setting up Amazon Mechanical Turk tasks. We acknowledge partial support from the National Science Foundation (NSF) award: CNS-1513721: ``Context-Aware Harassment Detection on Social Media'', the National Institute on Drug Abuse (NIDA) Grant No. 5R01DA039454-02: ``Trending: Social Media Analysis to Monitor Cannabis and Synthetic Cannabinoid Use'' and the National Institutes of Mental Health (NIMH) award: 1R01MH105384-01A1: ``Modeling Social Behavior for Healthcare Utilization in Depression''. Points of view or opinions in this document are those of the authors and do not necessarily represent the official position or policies of the NSF, NIDA, or NIMH.

\bibliographystyle{aaai}

\end{document}